\def\year{2020}\relax
\documentclass[letterpaper]{article} 
\usepackage{aaai20}  
\usepackage{times}  
\usepackage{helvet} 
\usepackage{courier}  
\usepackage[hyphens]{url}  
\usepackage{graphicx} 
\usepackage{multirow}
\usepackage{adjustbox}
\usepackage{esdiff}
\usepackage{amsmath}
\usepackage{bbold}
\usepackage{algorithm}  
\usepackage{algorithmicx}  
\usepackage{algpseudocode}  
\title{Effective AER Object Classification Using Segmented Probability-Maximization Learning in Spiking Neural Networks }

\author{Qianhui Liu,\textsuperscript{\rm 1} 
	Haibo Ruan,\textsuperscript{\rm 1} 
	Dong Xing,\textsuperscript{\rm 1}
	Huajin Tang,\textsuperscript{\rm 1,2} 
	Gang Pan\textsuperscript{\rm 1,2}\thanks{Corresponding author}\\
	\textsuperscript{\rm 1}College of Computer Science and Techology, Zhejiang University, Hangzhou, China\\ 
	\textsuperscript{\rm 2}Zhejiang Lab, Hangzhou, China\\
	\{qianhuiliu, hbruan, dongxing, htang, gpan\}@zju.edu.cn
 }
 
\begin{document}

\maketitle

\begin{abstract}
	Address event representation (AER) cameras have recently attracted more attention due to the advantages of high temporal resolution and low power consumption, compared with traditional frame-based cameras. Since AER cameras record the visual input as asynchronous  discrete events, they are inherently suitable to coordinate with the spiking neural network (SNN), which is biologically plausible and energy-efficient on neuromorphic hardware. However, using SNN to perform the AER object classification is still challenging, due to the lack of effective learning algorithms for this new representation. To tackle this issue, we propose an AER object classification model using a novel \textit{segmented probability-maximization} (SPA) learning algorithm. Technically, 1) the SPA learning algorithm iteratively maximizes the probability of the classes that samples belong to, in order to improve the reliability of neuron responses and effectiveness of learning; 2) a \textit{peak detection} (PD) mechanism is introduced in SPA to locate informative time points segment by segment, based on which information within the whole event stream can be fully utilized by the learning. Extensive experimental results show that, compared to state-of-the-art methods, not only our model is more effective, but also it requires less information to reach a certain level of accuracy.
\end{abstract}

\section{Introduction}
Address event representation (AER) cameras are neuromorphic devices imitating the mechanism of human retina. Contrary to traditional frame-based cameras, which record the visual input from all pixels as images at a fix rate, with AER cameras, each pixel individually emits events when it monitors sufficient changes of light intensity in its receptive field.  
The final output of the camera is a stream of events collected from each pixel, forming  an asynchronous and sparse representation of the scene.
AER cameras naturally respond to moving objects and ignore static redundant information, resulting in significant reduction of memory usage and energy consumption. 
Moreover, AER cameras capture visual information at a significantly higher temporal resolution than traditional frame-based cameras, achieving accuracy down to sub-microsecond levels under optimal conditions \cite{orchard2015hfirst}.
Commonly used AER cameras include the asynchronous time-based image sensor (ATIS) \cite{posch2011qvga}, dynamic vision sensor (DVS) \cite{lichtsteiner2008128,lenero20113}, dynamic and active pixel vision sensor (DAVIS) \cite{brandli2014240}.
 
 This event-based representation is inherently suitable to coordinate with the spiking neural network (SNN) since SNN also has the event-based property. SNN is generally more solid on biological plausibility and more powerful on processing both spatial and temporal information. It may also very help to build cyborg intelligent systems \cite{wu2013cyborg,wu2014cyborg}.
Moreover, SNN has the advantage of energy efficiency, for example, current implementations of SNN on neuromorphic hardware use only a few nJ or even pJ for transmitting a spike \cite{diehl2015unsupervised}.

However, the novelty of this representation also poses several challenges to AER object classification using SNN.
Firstly, the event streams from AER cameras are not stable, compared with the video streams from tradition cameras. 
AER cameras are sensitive to the dynamic information within the visual receptive field.  
Along with the events relevant to the objects, factors like camera shaking and subtle changes of environmental light will generate a large quantity of noisy events, which will impact the reliability of neuron responses and the learning performance of SNN.
 Secondly, the event stream from AER camera records massive information of a period of time, and a mechanism is required to make full use of the information for training and reach a competitive level of accuracy despite the testing information is not complete.
We will make steps towards improving the effectiveness of classification using SNN.

We propose an AER object classification model, which consists of an event-based spatio-temporal feature extraction and a new \textit{segmented probability-maximization} (SPA) learning algorithm of SNN.
Firstly, the feature extraction obtains the representative features of the output of AER cameras, reducing the effect of noisy events to some extent and maintaining the precise timing of the output events.
Feature extraction employs the spiking neurons to utilize the precise timing information inherently present in the output of AER cameras and uses spike timing to represent the spatio-temporal features \cite{orchard2015hfirst}. 
Then, the SPA supervised learning algorithm constructs an objective function based on the probability of the classes that samples belong to and iteratively updates the synaptic weights with gradient-based optimization of the objective function.
To fully utilize the massive information in the event stream covering a period of time, we introduce a \textit{peak detection} (PD) in SPA to trigger the weight updating procedure based on the informative time points located segment by segment.
The SPA learning algorithm enables the trained neurons to respond actively to their representing classes. Therefore, the classification decision can be determined by the firing rate of every trained neuron. 
We perform extensive experiments to verify our model, and the results show that our model is more effective when compared with state-of-the-art methods, and requires less information to reach a certain level of accuracy.

\section{Related Work}
\subsection{Event-Based Features and Object Classification}
\cite{perez2013mapping} presented a methodology by training a frame-driven convolutional neural network (ConvNet) with images (frames) by collecting events during fixed time intervals and mapping the frame-driven ConvNet to an event-driven ConvNet.
\cite{neil2016effective} introduced various deep network architectures including a deep fusion network composed of convolutional neural networks and recurrent neural networks to jointly solve recognition tasks.
Different from the existing deep learning based methods, \cite{lagorce2017hots} proposed the spatio-temporal features based on recent temporal activity within a local spatial neighborhood called time-surfaces and a hierarchy of event-based  time-surfaces for pattern recognition (HOTS). \cite{sironi2018hats} introduced  the Histograms of Averaged Time Surfaces (HATS) for feature representation of event-based object recognition.
In addition, there are some existing works using SNN for event-based features and object classification.  \cite{zhao2015feedforward} presented an event-driven HMAX network for feature extraction and a tempotron classifier of SNN for classification.  Further, \cite{orchard2015hfirst} proposed an HMAX inspired SNN for object recognition (HFirst). HFirst does not require extra coding and consistently uses precise timing of spikes for feature representation and learning process. 
\cite{cohen2016skimming} presented an implementation of Synaptic Kernel Inverse Method (SKIM), which is a learning method based on principles of dendritic computation, in order to perform a large-scale AER object classification task.
\cite{liu2019unsupervised} proposed a multiscale spatio-temporal feature (MuST) representation of AER events and an unsupervised rocognition approach.

\subsection{SNN Learning Algorithm}
SpikeProp \cite{bohte2002error} is one of the most classical SNN learning algorithm. It constructs an error function by the difference between the desired and actual output spikes, then updates the synaptic weights based on gradient descent. Other learning algorithms that also define the desired spike sequences are ReSuMe \cite{ponulak2010supervised}, SPAN \cite{mohemmed2012span}, PSD \cite{yu2013precise}, etc. Recently, membrane voltage-driven methods have emerged in an attempt to improve the learning efficiency and accuracy of spiking neurons.
MPD-AL \cite{zhang2019mpd} proposed a membrane-potential driven aggregate-label learning algorithm, which constructs an error function based on the membrane potential trace and the fixed firing threshold of the neuron. It dynamically determines the number of desired output spikes instead of enforcing a fixed number of desired spikes. However, these algorithms need to output a corresponding spike sequence for classification. Note that, for the AER object classification task, it is desired to give a result in time instead of waiting until the output sequence has been fully generated. 

Tempotron \cite{gutig2006tempotron,qi2018jointly} is also a voltage-driven learning algorithm and aims to train the output neuron to fire a single spike or not according to its class label. If the neuron is supposed to fire (or not fire, on the other hand) but it actually fails to do so (or does fire, vice versa), then the weights should be modified. Tempotron implements a gradient descent dynamics that minimizes the error defined as the difference between the maximal membrane potential and the firing threshold. This kind of ``single spike" classifier tends to be affected by noise and thus is not suitable for the task of  AER object classification.

The proposed SPA learning algorithm for AER object classification aims to enable the trained neurons to respond actively to their representing classes. The model gives the classification decision by choosing the class with the highest average firing rate. In this way, we do not need to wait for the output sequence to complete and can directly give the results based on the firing rates at the current time.  Therefore, our proposed SPA algorithm is more robust and flexible for AER object classification task.
\begin{figure*}[!t]	
	\centering
\includegraphics[width=.85\textwidth]{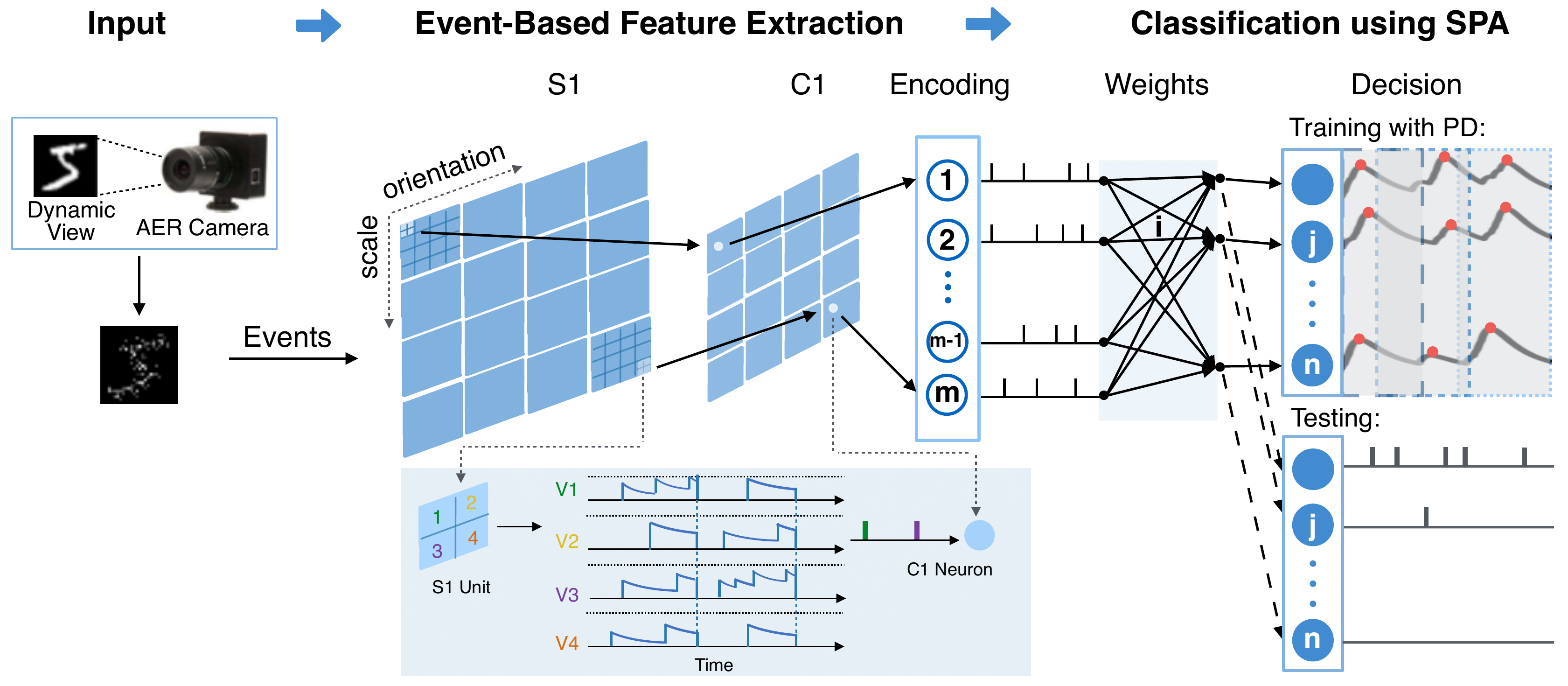}
	\caption{The flow chart of the proposed AER object classification.  The events from the AER camera are firstly sent to $S1$ layer. Neurons in $S1$ layer have their own receptive scale and respond best to a certain orientation. Neurons of the same receptive scale and orientation are organized into one feature map (denoted by blue squares). $S1$ feature maps are divided into adjacent non-overlapping $2 \times 2$ cell regions, namely $S1$ units. If the membrane voltage of $S1$ neuron exceeds its threshold, the neuron will fire a spike to the $C1$ layer and all neurons in the same unit are reset to $V_{reset}$. 
	$C1$ neurons then transmit the feature spikes to the encoding layer.
		 During training, the weights of synapses from  encoding neurons to decision neurons are updated by the proposed SPA learning algorithm, which locates the informative time points $t_{peak}$ segment by segment with PD and maximizes the corresponding probability based on the $t_{peak}$. The gray bands and red dots denote the segments and informative time points respectively. After training is done, the synaptic weights are fixed.
		During testing, the final classification decision is determined by averaging the firing rates of decision neurons per class and choosing the class with the highest average firing rate. Note that we separately show the decision layer during training and testing  because of the different readout way during training and testing.}
	
	\label{fig:framework}
	
\end{figure*}

\section{Method}
In this section, we will introduce the proposed AER object classification model, which mainly consists of an event-based spatio-temporal feature extraction and a novel SNN learning algorithm. The flow chart of the model is shown in Figure \ref{fig:framework}. 
\subsection{Event-Based Feature Extraction}
\begin{figure}[!t]
	\centering
		\includegraphics[width=.78\columnwidth]{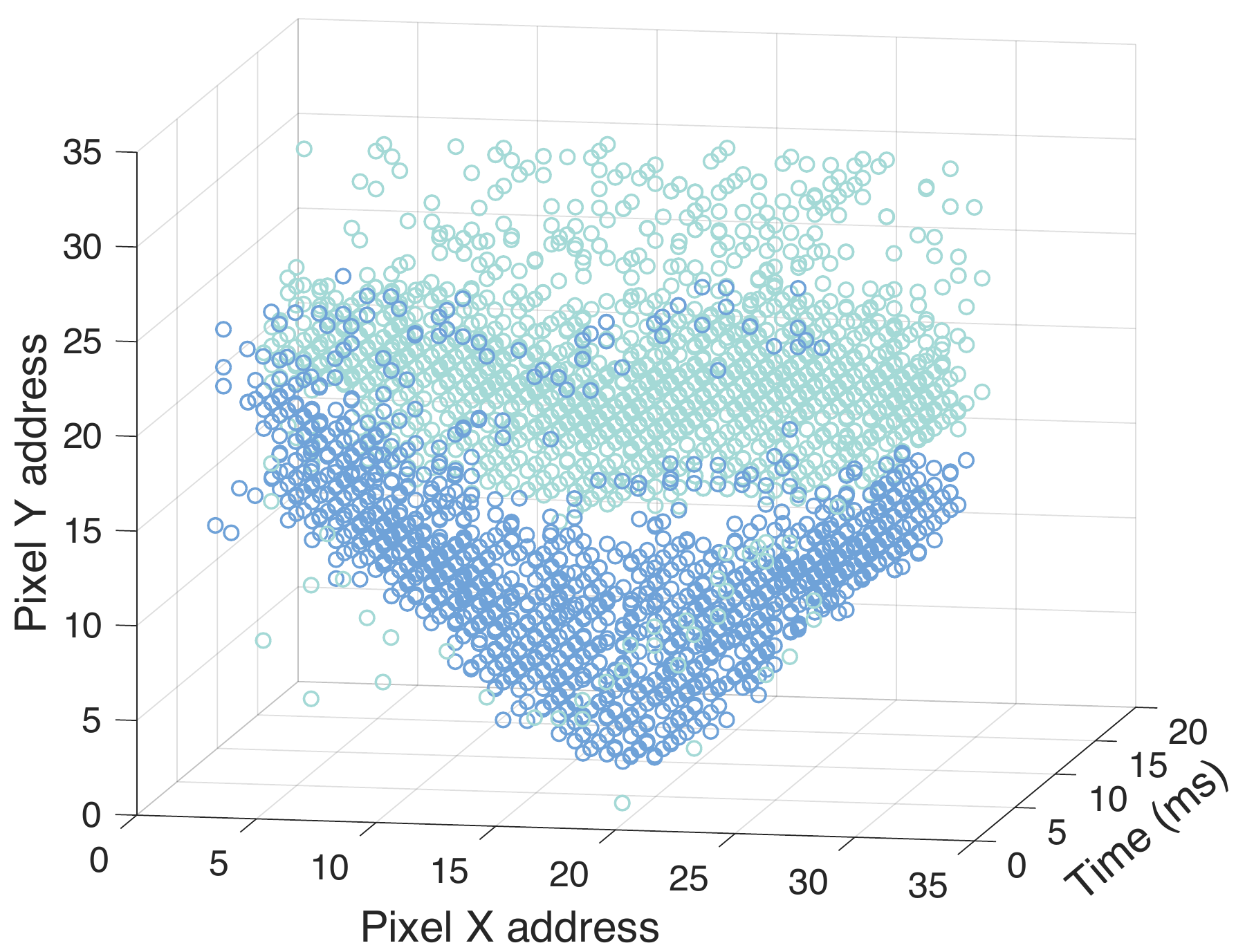}
	\caption{Visualization of the AER event stream representing the ``heart" symbol in Cards dataset \cite{serrano2015poker}. ON and OFF events are represented by cyan and blue circles respectively.  }
	\label{fig:AERevent}
\end{figure}

Feature extraction follows the manner of Hierarchical Model and X (HMAX) \cite{riesenhuber1999hierarchical}, a popular bio-inspired model mimicking the information processing of visual cortex. We employ a hierarchy of $S1$ layer and $C1$ layer, corresponding to the simple and complex cells in primary visual cortex V1 respectively. The simple cells combine the input with a bell-shaped tuning function to increase feature selectivity and the complex cells perform the maximum operation to increase feature invariance \cite{riesenhuber1999hierarchical}. We model the simple and complex cells using spiking neurons which generate spikes to represent the features of the output events of AER cameras. 
Given an AER camera with pixel grid size $N \times M$, the $i$-th event is described as: 
\begin{equation}\label{event1}
e_i
= [e_{x_i}, e_{y_i},e_{t_i},e_{p_i}], i \in \{1, 2, \dots, I\}
\end{equation}
where $(e_{x_i},e_{y_i}) \in \{1,2,\dots,N\} \times \{1,2,\dots,M\}$ is the position of the pixel generating the $i$-th event, $e_{t_i} \geq 0$ the timestamp at which the event is generated, $e_{p_i} \in \{-1, 1\}$ the polarity of the event, with $-1, 1$ meaning respectively OFF and ON events, and $I$ the number of events. Figure \ref{fig:AERevent} shows the visualization of an AER event stream representing the ``heart" symbol in Cards dataset \cite{serrano2015poker}.

\subsubsection{From Input to $S1$ Layer}The output events of the AER camera are sent as input to the $S1$ layer, in which each event is convolved with a group of Gabor filters \cite{zhao2015feedforward}. Each filter models a neuron cell that has a certain scale $s$ of receptive field and responds best to a certain orientation $\theta$. The function of Gabor filter can be described with the following equation:
\begin{equation}\label{Gabor1}
\begin{aligned}
G(\Delta x,\Delta y)
& = \exp(-\frac{X^2+\gamma ^2 Y^2}{2\sigma ^2})\cos(\frac{2\pi}{\lambda}X) \\
X &= \Delta x \cos\theta + \Delta y\sin\theta  \\
Y &= -\Delta x\sin\theta + \Delta y\cos\theta 
\end{aligned}
\end{equation}
where  $\Delta x$ and $\Delta y$ are the spatial offsets between the pixel position $(x,y)$ and the event address $(e_x, e_y)$, $\gamma$ is the aspect ratio.  The wavelength $\lambda$ and effective width $\sigma$ are parameters determined by scale $s$. 
We choose four orientations ($0^{\circ}$, $45^{\circ}$, $90^{\circ}$, $135^{\circ}$) and a range of sizes from $3\times 3$ to $9\times9$ pixels with strides of two pixels for Gabor filters.
Other parameters in the Gabor filter are inherited from \cite{zhao2015feedforward}. 


Each spiking neuron corresponds to one pixel in the camera and neurons of the same receptive scale and orientation are organized into one feature map.
The  membrane voltages of neurons in feature maps are initialized as zeros, then updated by adding each element of the filters to the maps at the position specified by the address of each event.  At the same time, the decay mechanism of the spiking neuron is maintained to eliminate the impact of very old events on the current response.
The membrane voltage of the neuron at position $(x,y)$ and time $t$ in the map of specific scale $s$ and orientation $\theta$ can be described as:
\begin{multline}\label{ResMap}
V(x,y,t;s,\theta) = 		
\sum_{i} 
\mathbb{1}\{x \in \mathcal{X}(e_{x_i})\}\mathbb{1}\{y \in \mathcal{Y}(e_{y_i})\}\\
G(x-e_{x_i},y-e_{y_i};s,\theta)\exp(-\frac{t-e_{t_i}}{\tau_{m}})
\end{multline}
where $\mathbb{1}\{\cdot\}$  is the indicator function,  $\mathcal{X}(e_{x_i}) = [e_{x_i} -s, e_{x_i} + s]$ and  $\mathcal{Y}(e_{y_i}) = [e_{y_i} -s, e_{y_i} + s]$ denote the receptive field of the neuron, and $\tau_{m}$ denotes the decay time constant. The function $G(\cdot; s, \theta)$ represents the filters grouped by $s$ and $\theta$.  When the membrane voltage of neuron in $S1$ layer exceeds its threshold $V_{thr}^{s_1}$, the neuron will fire a spike. The threshold $V_{thr}^{s_1}$ is set as $2$ in this paper.
\subsubsection{From $S1$ Layer to $C1$ Layer}
Feature maps in $S1$ layer are divided into adjacent non-overlapping $2 \times 2$ cell regions, namely $S1$ units. When a neuron in $S1$ layer emits a spike, other neurons in the same unit will be inhibited, and all neurons in this unit will be forced to reset to $V_{reset}$, which is typically set as $0$.  
This lateral inhibition mechanism ensures that only maximal responses in $S1$ units can be propagated to the subsequent layers.
Therefore, we only need to observe which neuron generates an output spike to the $C1$ neuron first instead of comparing the neuron responses. 

\subsection{Classification Using Segmented Probability-Maximization (SPA) Learning Algorithm}

After extracting the spatio-temporal features, we introduce the SPA learning algorithm of SNN to enable the trained neurons to respond actively to their representing classes. During classification, the model will give the result based on the firing rate of each trained neuron.

\subsubsection{Neuron Model}
In this paper, we employ the  Leaky Integrate-and-Fire (LIF) model, which has been used widely to simulate spiking neurons and is good at processing temporal information, to describe the neural dynamics. In LIF model, each incoming spike induces a postsynaptic potential (PSP) to the neuron.
For an incoming spike received at $t_i$, the normalized PSP kernel is defined as follows:
\begin{equation}\label{psp}
K(t-t_i) = V_0(\exp(\frac{-(t-t_i)}{\tau_m})-\exp(\frac{-(t-t_i)}{\tau_s}))
\end{equation}
where $\tau_m$ and $\tau_s$ denote decay time constants of membrane integration and synaptic currents respectively,  and the ratio between them is fixed at $\tau_m/ \tau_s = 4$. The coefficient $V_0$ normalizes PSP so that the maximum value of the kernel is 1. The membrane voltage $V(t)$ of the decision neuron is described as:
\begin{equation}\label{lif}
V(t) = \sum_{i} w_i  \sum_{t_i} K(t-t_i) +V_{rest} 
\end{equation}
where $w_i$ and $t_i$ are the synaptic weight and the firing time of the afferent $i$. $V_{rest}$ denotes the resting potential of the neuron, which is typically set as $0$. 

\subsubsection{SPA Learning Algorithm}
We adopt a biologically-inspired activation function called \textit{Noisy Softplus}, which 
is well-matched to the response function of LIF spiking neurons \cite{liu2016noisy}.  
The mean input current in \cite{liu2016noisy} can be approximated to the effective input current, which can be reflected by 
peak voltage $V(t_{peak})$ of the receptive neuron. Here $t_{peak}$ denotes the time at which the postsynaptic voltage reaches its peak value. 
In the following, we use $V_{peak}$ to represent $V(t_{peak})$ for short.
We rewrite the\textit{ Noisy Softplus} function in \cite{liu2016noisy} as:
\begin{equation}\label{f1}
f_{out} = \log(\exp( {V_{peak}} )+1)
\end{equation}
where $f_{out}$ is the normalized output firing rate. For a $C$-class categorization task, we need $C$ decision neurons, each representing one class.  The output firing rate of the neuron representing class $j$ is:
\begin{equation}\label{f2}
f_{out}^{j} =\log(\exp( {V^{j}_{peak}} )+1)
\end{equation}
where  $V_{peak}^{j}$ is the peak voltage of the neuron representing class $j$.

The classification decisions are made based on the firing rates of spiking neurons in decision layer, so the aim of the learning is to train the decision neurons to respond actively to the input patterns of the class they represent.
To improve the reliability of neuron responses and effectiveness of learning,  
we introduce the probability of the class that the sample actually belongs to, and constantly increase the probability.
We define the probability that $k$-th sample belongs to class $j$ as:
\begin{equation}\label{p1}
P(\hat{c}_k = j) = \frac{f_{out}^{j}}{\sum_{j'=1}^{C}f_{out}^{j'}}
\end{equation}
where $\hat{c}_k$ denotes the predicted class of the $k$-th sample. 
We will use $f_{sum}$ to denote ${\sum_{j'=1}^{C}f_{out}^{j'}}$  for convenience.
Furthermore, we use the cross-entropy to define the loss function of $k$-th sample as:
\begin{equation}\label{l1}
L_k = - \log\left(P(\hat{c}_k = c_k)\right)
\end{equation}
where ${c}_k$ denotes the actual class of the $k$-th sample.

We then minimize the cost function with gradient descent optimization and iteratively update the synaptic weight $w_i$ by:
\begin{equation}\label{w1}
\Delta w_i  = - \lambda \frac{\partial L_k}{\partial V_{peak}^{j}}\frac{\partial V^{j}_{peak}}{\partial w_i}
\end{equation}
where $\lambda$ is the learning rate which is set as $0.1$, and $\frac{\partial L_k}{\partial V_{peak}^{j}}$ and $\frac{\partial V^{j}_{peak}}{\partial w_i}$ are evaluated as below:
\begin{equation}\label{w13}
\frac{\partial L_k}{\partial V_{peak}^{j}} = 
\left\{ \begin{array}{@{}ll@{}}
-(f'_{out})^j\frac{f_{sum}-f_{out}^j}{f_{sum}f_{out}^j} 
& j = c_k\\
 (f'_{out})^j\frac{1}{f_{sum}}
& j\not=c_k\\
\end{array}\right.
\end{equation}
\begin{equation}\label{w23}
\frac{\partial V^{j}_{peak}}{\partial w_i} = 
 \sum\limits_{t_S \le t_i<t_{peak}^{j}} K(t_{peak}^{j} - t_i)
\end{equation}
where $t_S$ is the starting time of an event stream.


\subsubsection{Peak Detection}
There may contain multiple voltage peaks of a neuron in an event stream covering a period of time.
The voltage peak is triggered by a burst of incoming feature spikes, indicating that the neuron has received a large amount of information at the time of the voltage peak.
To further utilize these informative time points,
we propose a \textit{peak detection} mechanism (PD) to locate the peaks of the membrane voltage segment by segment.
The principle of PD is as follows: within a search range with length $t_{R}$ starting at $t_{S}$, 
$t_{peak}$ of  the  neuron representing class $j$ is defined as:
\begin{equation}\label{pd1}
t_{peak}^j = \arg\max\limits_t \left\{ V^j (t)| t \in  (t_{S},t_{S} +t_{R}]   \right\}
\end{equation}
where $V^j(\cdot)$ is the membrane voltage of the neuron representing class $j$.
If multiple time points in the current segment meet the criterion, the earliest one is chosen. After locating the voltage peak in the current segment for each neuron, $t_{S}$ will be updated as:
\begin{equation}\label{pd2}
t_{S} = \max_{1\le j \le C} t_{peak}^{j}
\end{equation}
The full procedure for SPA learning algorithm with PD  is summarized in Algorithm 1. 
\begin{algorithm}[!b]  
	\caption{SPA  learning algorithm for classification}  
	\begin{algorithmic}[1]  		
		\Require $FeatureSpikes$ from encoding layer
		\Ensure Synaptic weight $w$
		\Function {SPA}{$FeatureSpikes$}
		\State Initialize the neuron membrane voltage $V$ and synaptic weights $w$. Set the iteration time $n_{iter}$, the time length $L$ and the learning rate $\lambda$;
		\While {$n_{iter}$ not reached}
		\State Calculate the membrane voltage $V$ by (\ref{lif});
		\State Initialize $t_S$ = 0;
		\While {$t_S < L $}  
		\State Find the $t^{peak}_j$ of each decision neuron $j$ in the search range of ($t_S$,  $t_S + t_{R}$] according to (\ref{pd1});
		\State Update  $w_i$ by $\Delta w_i$ for each afferent $i$ by (\ref{w1});	
		\State Update $t_S$ according to (\ref{pd2});  
		\EndWhile 
		\EndWhile 
		\State Return $w$;
		\EndFunction
		\label{code:recentEnd}  
	\end{algorithmic}  
\end{algorithm}  
\begin{figure*}[!t]
	\centering
	\includegraphics[width=.85\textwidth]{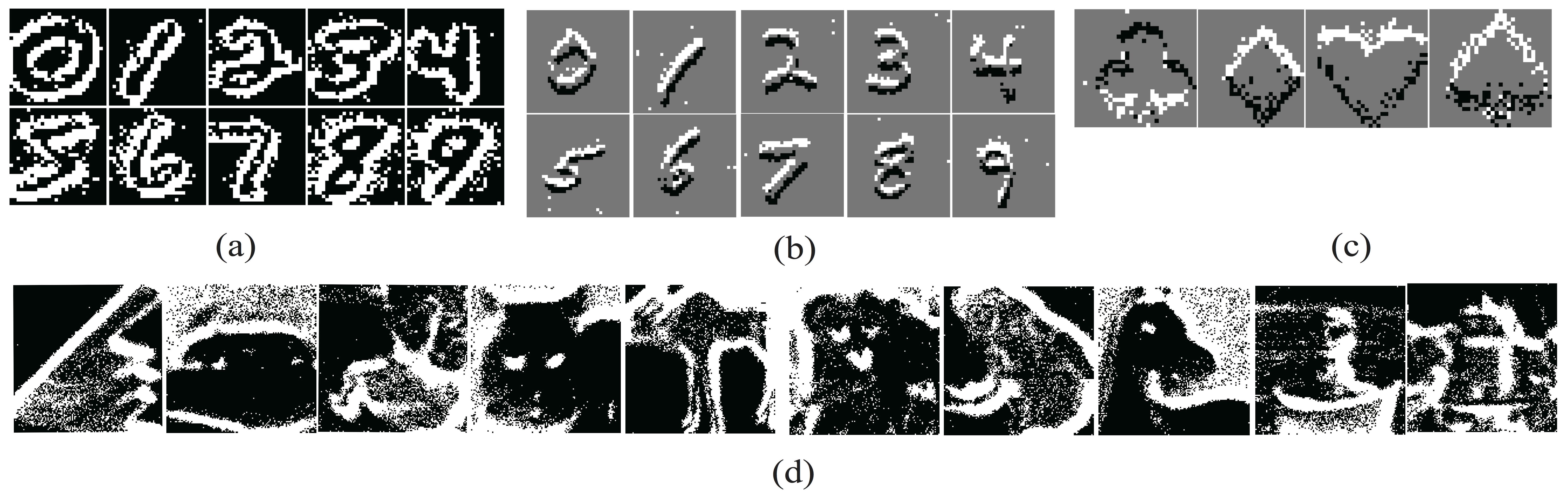}
	\caption{Some reconstructed images from the used datasets: (a) MNIST-DVS dataset; (b) NMNIST dataset; (c) Cards dataset; (d) CIFAR10-DVS dataset. }
	\label{fig:aerdata}
\end{figure*}
\subsubsection{Network Design}  We use the extracted feature spikes from $C1$ layer as the input of the classification process. Encoding neurons are fully connected to the decision neurons. Since the activity of one single neuron can be easily affected, the population coding is adopted to improve the reliability of information coding \cite{natarajan2008encoding}. In this paper, each input class is set to associate with a population of $10$ decision neurons.

During training, the synaptic weights are firstly initialized with random values and updated using SPA learning algorithm. 
When training is done, we keep the synaptic weights fixed, and set the threshold of decision neurons $V^{dec}_{thr}$  as $1$. During testing, 
when the decision neuron's membrane potential is higher than its threshold $V^{dec}_{thr}$, the neuron will fire a spike and its membrane potential will be reset to $V_{reset}$. 
The predicted class for the input is determined by averaging the firing rates of neurons per class and then choosing the class with the highest average firing rate.

\section{Experimental Results}
In this section, we evaluate the performance of our proposed approach on several public AER datasets and compare it with other AER classification methods.

\subsection{Datasets}
Four publicly available datasets are used to analyze the performance.

1) \textbf{MNIST-DVS} dataset \cite{lichtsteiner2008128}: it is obtained with a DVS by recording 10,000 original handwritten images in MNIST moving with slow motion. 

2) \textbf{ Neuromorphic-MNIST (NMNIST)}  dataset \cite{orchard2015converting}: it is obtained by moving an ATIS camera in front of the original MNIST images. It consists of 60,000 training and 10,000 testing samples. 

3) \textbf{Cards} dataset \cite{serrano2015poker}: 
it is captured by browsing a card deck in front of the sensitive DVS camera and recording the information in an event stream. The event stream consists of 10 cards for each of the 4 card types (spades, hearts, diamonds, and clubs).

4) \textbf{CIFAR10-DVS} dataset \cite{li2017cifar10}: it consists of 10,000 samples, which are obtained by displaying the moving CIFAR10 images on a monitor and recorded with a fixed DVS camera. 

Figure \ref{fig:aerdata} shows some samples of these four datasets.

\subsection{Experimental Settings} 
We randomly partition the used datasets into two parts for training and testing. The result is obtained over multiple runs with different training and testing data partitions.
For fair comparison, the results of competing methods and ours are obtained under the same experimental settings. The results of competing  methods are from the original papers \cite{orchard2015hfirst,lagorce2017hots,zhao2015feedforward}, or (if not in the papers) from the experiments using the code with our optimization.

\subsection{Performance on Different AER Datasets}
\begin{table*}[!t]
	\caption{Comparison of classification accuracy on four datasets.}
	\begin{center}		
		\begin{tabular}{lcccc}
			\hline
			\hline
			&\textbf{MNIST-DVS}&\textbf{NMNIST}& \textbf{Cards}  & \textbf{CIFAR10-DVS}  \\
			\hline
			Zhao's \cite{zhao2015feedforward}& 88.1\%&85.6\%&86.5\% &21.9\%\\
			HFirst \cite{orchard2015hfirst} & 78.1\%& 71.2\%&97.5\%  & 7.7\% \\
			HOTS \cite{lagorce2017hots}&80.3\%&80.8\%& 100.0\% & 27.1\%\\
			\hline
			This Work&96.7\%&96.3\%&   100.0\% &32.2\%\\
			\hline
			\hline 
		\end{tabular}
	\end{center}
	\label{table:comparison}
\end{table*}
\subsubsection{On MNIST-DVS Dataset}
The time length of reconstructing a digit silhouette in this dataset is no more than $80 ms$, thus the search range $t_{R}$  and the parameter $\tau_m$ are both set as $80 ms$.
This dataset has 10,000 samples, 90\% of which are randomly selected for training and the remaining ones are used for testing. The performance is averaged over 10 runs. 

Our model gets the classification accuracy of 96.7\%, or 3.3\% error rate on average. Table \ref{table:comparison} shows that our model gets a higher performance and in addition achieves an more than 3 times smaller error rate than Zhao's method (11.9\%),  HFirst (21.9\%), and HOTS (19.7\%)  \cite{zhao2015feedforward,orchard2015hfirst,lagorce2017hots}.

\subsubsection{On NMNIST Dataset}
This dataset records the event streams produced by 3 saccadic movements of the DVS camera. The time length of each saccadic movement is about $120 ms$, thus the search range $t_{R}$ and the parameter $\tau_m$ are set as $120 ms$.
This dataset is inherited from MNIST, and has been partitioned into 60,000 training samples and 10,000 testing samples by default.

Our model gets the classification accuracy of 96.3\% on average. Table \ref{table:comparison} shows our model outperforms Zhao's method \cite{zhao2015feedforward},  HFirst \cite{orchard2015hfirst}, and HOTS \cite{lagorce2017hots} by a margin of 10.7\%,  25.1\% and 15.5\% respectively. Notice that HFirst has relatively poor performance on NMNIST, compared with the other methods.
However, this drop in accuracy is expected because HFirst is designed to detect simple objects, while great variation of object appearance exists in the NMNIST dataset.
In addition, SKIM network on NMNIST dataset \cite{cohen2016skimming} achieves an accuracy of 92.9\%, which is also 3.8\% less than our model.

\subsubsection{On Cards Dataset}
The time length of the recordings in this dataset is about $20 ms$. Since  $8 ms$ is enough to reconstruct a card silhouette, we set the search range $t_{R}$ and $\tau_m$ as $8 ms$.
For each category of this dataset, 50\% are randomly selected for training and the others are used for testing. The performance is averaged over 10 runs to get the result.

Our model achieves the classification accuracy of 100\% for the testing set. HFirst and HOTS also reach relatively high accuracy, while Zhao's model only gets an accuracy of 86.5\%. This is because the number of training samples in this dataset is very limited (only 5 samples per class), and at most one spike in Zhao's method is emitted for each encoding neuron to represent features.
Therefore, there is not enough effective information for tempotron classifier to train the SNN.
\begin{figure}[!t]
	\centering
		\includegraphics[width=.9\columnwidth]{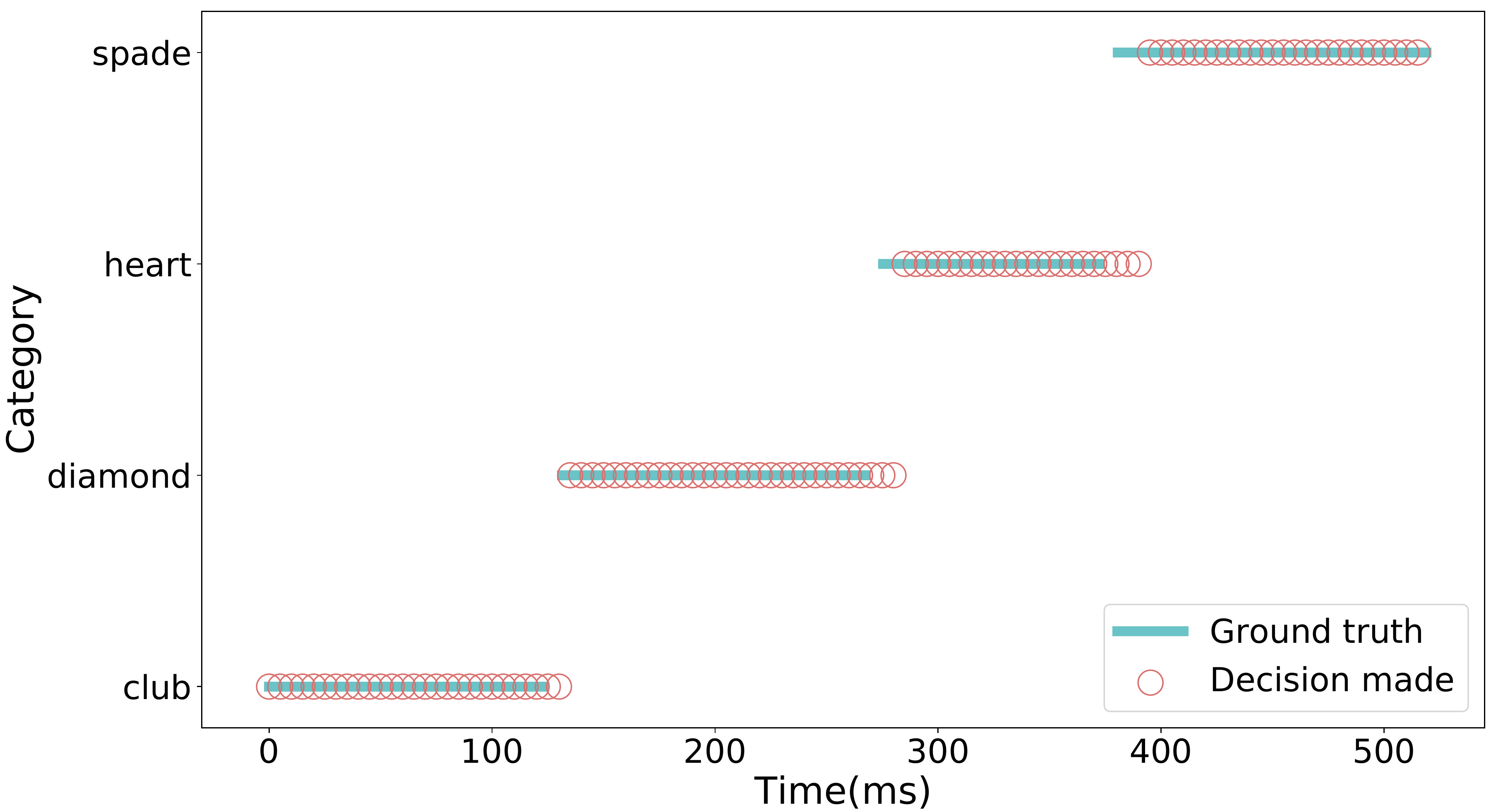}
		
	\caption{Performance of the proposed model on a continuous event stream from Cards dataset. All testing symbols are connected one by one into a continuous event stream and then fed to the model for evaluation. The cyan lines represent the ground truth of classification, and the red circles denote the decisions made by our model every $5 ms$. }
	\label{fig:cardpips}
\end{figure}

We further run our model on a continuous event stream which combines all the testing samples, since this dataset is originally in the  form of a stream of events. The result is shown in Figure \ref{fig:cardpips}. Every $5 ms$ we give a decision (one red circle in the figure). We can see that, at the beginning of the appearance of a new type, the decisions we made are slightly delayed. This is because, when a new type appears, the neurons representing the new class have not accumulated enough responses to outperform the neurons representing the former class. Nevertheless, after about $15ms$, the decisions can match very well with the ground truth.

\subsubsection{On CIFAR10-DVS  Dataset}
In this dataset, $80 ms$ is enough to reconstruct a object silhouette, thus the search range $t_{R}$  and the parameter $\tau_m$ are both set as $80 ms$. We randomly select 90\% of samples for training and the others for testing. The experiments are repeated 10 times to obtain the average performance. 

The classification task of this dataset is more challenging because of the complicated object appearance and the large intra-class variance, therefore the classification results on this dataset of all the compared methods are relatively poor. Nevertheless, our model achieves the classification accuracy of 32.2\%, which is still higher than other methods.

\subsection{Effects of the SPA} 
In this section, we carry out more experiments to demonstrate the effects of our model using SPA in detail. The experiments are conducted on MNIST-DVS dataset and the parameter settings are the same as the previous section.

\subsubsection{Sample Efficiency} 
Sample efficiency measures the quantity of samples or information required for a model to reach a certain level of accuracy. 
 In the AER object classification task, the length of the event stream determines the amount of information.
We examine the impact of the time length on the algorithm.
The experiments are conducted on recordings with the first $100 ms$, $200 ms$, $500 ms$ and full length (about $2 s$) of the original samples, respectively. Since Zhao's method achieves a competitive classification result on the full length of MNIST-DVS as shown in Table \ref{table:comparison}, we list the results of both Zhao's method and ours in Table \ref{table:mnistdvs}. 

It can be noticed that: 1) the accuracy of both two methods keeps increasing when longer recordings are used, which is because longer recordings provide more information; 2) our model consistently outperforms Zhao's method on recordings with every time length in Table \ref{table:mnistdvs}. In fact, even on the recordings with time length $100ms$, our model still yields a relatively better result than Zhao's method on the recordings of full length. 
This result demonstrates that with the same or even less information,  our model could reach a better classification accuracy, which proves its sample efficiency.
\begin{table}[h]
	\caption{Performance on recordings with different time length of MNIST-DVS dataset.}
	\begin{center}		
		\begin{tabular}{ccc}
			\hline
			\hline
		    Time Length & {Zhao's}&  {This Work}\\
			\hline
			100 ms  &76.9\%& 89.4\% \\
			200 ms &82.6\%& 92.7\%  \\
			500 ms &85.9\%& 94.9\% \\
			Full (about 2s) &88.1\%& 96.7\%\\
			\hline
			\hline 
		\end{tabular}
	\end{center}
	\label{table:mnistdvs}
\end{table}
\subsubsection{Inference with Incomplete Information}
\begin{figure}[!t]
	\centering

		\includegraphics[width=.92\columnwidth]{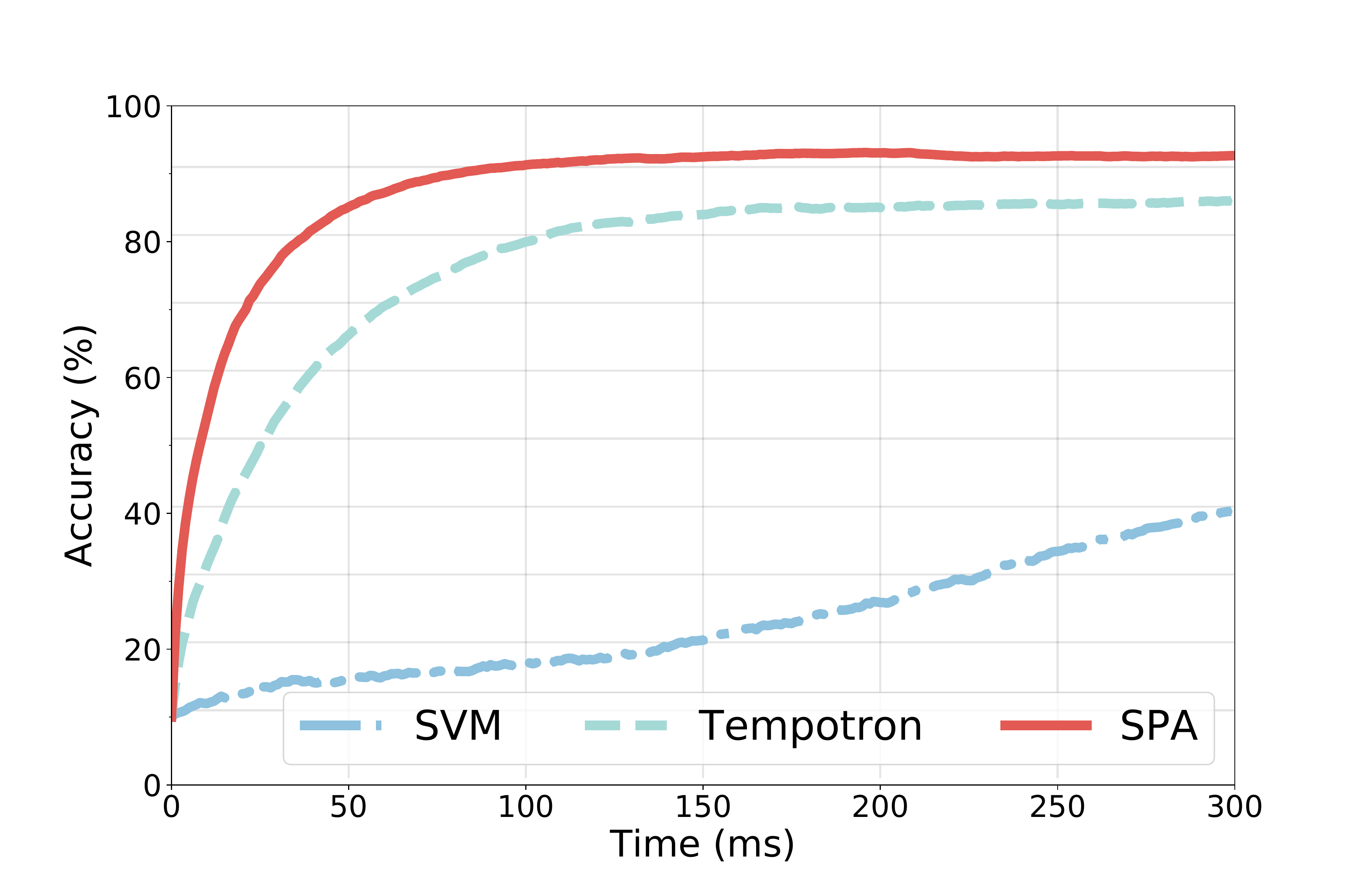}
	\caption{Performance of the inference with incomplete information on MNIST-DVS dataset. }
	\label{fig:mnistcurve}
\end{figure}

Inference with incomplete information requires the model to be capable of responding with a certain level of accuracy, when information of the object is incomplete during testing.
We use the recordings of $500ms$ for training and observe the performance within the first $300ms$ recordings of three methods, including the proposed SPA learning algorithm, the tempotron learning algorithm used in Zhao's method, and the nontemporal classifier SVM (with the same feature extraction procedure). 
The results are averaged over 10 runs and shown in Figure \ref{fig:mnistcurve}. 

As the event stream flows in, the classification accuracy of models with the three algorithms keeps increasing. The model with SPA has the highest performance among all the methods, especially within the first $100ms$ when the input information is extremely incomplete. 
This is because in our SPA learning algorithm, each sample is trained several times based on the informative time points in every segment, which increases the diversity of the training information. Therefore, the model has a better generalization ability and can be promoted rapidly at the early stage, even though the information is incomplete.

\section{Conclusion}
In this paper, we propose an effective AER object classification model using a novel SPA learning algorithm of SNN. 
The SPA learning algorithm iteratively updates the weight by maximizing the probability of the actual class to which the sample belongs.
A PD mechanism is introduced in SPA to locate informative time points segment by segment on the temporal axis, based on which the information can be fully utilized by the learning. Experimental results show that our approach yields better performance on four public AER datasets, compared with other benchmark methods specifically designed for AER tasks. Moreover, experimental results also demonstrate the advantage of sample-efficiency and the ability of inference with incomplete information of our model. 

\section{Acknowledgments}
This work is partly supported by National Key Research and Development Program of China (2017YFB1002503), Zhejiang Lab (No. 2019KC0AD02), Ten Thousand Talent Program of Zhejiang Province (No.2018R52039), and National Science Fund for Distinguished Young Scholars (No. 61925603).

\bibliographystyle{aaai}
\bibliography{6583-cite}

\begin{thebibliography}{}

\bibitem[\protect\citeauthoryear{Bohte, Kok, and
  La~Poutre}{2002}]{bohte2002error}
Bohte, S.~M.; Kok, J.~N.; and La~Poutre, H.
\newblock 2002.
\newblock Error-backpropagation in temporally encoded networks of spiking
  neurons.
\newblock {\em Neurocomputing} 48(1-4):17--37.

\bibitem[\protect\citeauthoryear{Brandli \bgroup et al\mbox.\egroup
  }{2014}]{brandli2014240}
Brandli, C.; Berner, R.; Yang, M.; Liu, S.-C.; and Delbruck, T.
\newblock 2014.
\newblock A 240$\times$ 180 130 db 3 $\mu$s latency global shutter
  spatiotemporal vision sensor.
\newblock {\em IEEE Journal of Solid-State Circuits} 49(10):2333--2341.

\bibitem[\protect\citeauthoryear{Cohen \bgroup et al\mbox.\egroup
  }{2016}]{cohen2016skimming}
Cohen, G.~K.; Orchard, G.; Leng, S.-H.; Tapson, J.; Benosman, R.~B.; and
  Van~Schaik, A.
\newblock 2016.
\newblock Skimming digits: neuromorphic classification of spike-encoded images.
\newblock {\em Frontiers in neuroscience} 10:184.

\bibitem[\protect\citeauthoryear{Diehl and Cook}{2015}]{diehl2015unsupervised}
Diehl, P.~U., and Cook, M.
\newblock 2015.
\newblock Unsupervised learning of digit recognition using
  spike-timing-dependent plasticity.
\newblock {\em Frontiers in Computational Neuroscience} 9:99.

\bibitem[\protect\citeauthoryear{G{\"u}tig and
  Sompolinsky}{2006}]{gutig2006tempotron}
G{\"u}tig, R., and Sompolinsky, H.
\newblock 2006.
\newblock The tempotron: a neuron that learns spike timing--based decisions.
\newblock {\em Nature neuroscience} 9(3):420.

\bibitem[\protect\citeauthoryear{Lagorce \bgroup et al\mbox.\egroup
  }{2017}]{lagorce2017hots}
Lagorce, X.; Orchard, G.; Galluppi, F.; Shi, B.~E.; and Benosman, R.~B.
\newblock 2017.
\newblock Hots: a hierarchy of event-based time-surfaces for pattern
  recognition.
\newblock {\em IEEE Transactions on Pattern Analysis and Machine Intelligence}
  39(7):1346--1359.

\bibitem[\protect\citeauthoryear{Le{\~n}ero-Bardallo, Serrano-Gotarredona, and
  Linares-Barranco}{2011}]{lenero20113}
Le{\~n}ero-Bardallo, J.~A.; Serrano-Gotarredona, T.; and Linares-Barranco, B.
\newblock 2011.
\newblock A 3.6$\mu$s latency asynchronous frame-free event-driven
  dynamic-vision-sensor.
\newblock {\em IEEE Journal of Solid-State Circuits} 46(6):1443--1455.

\bibitem[\protect\citeauthoryear{Li \bgroup et al\mbox.\egroup
  }{2017}]{li2017cifar10}
Li, H.; Liu, H.; Ji, X.; Li, G.; and Shi, L.
\newblock 2017.
\newblock Cifar10-dvs: An event-stream dataset for object classification.
\newblock {\em Frontiers in Neuroscience} 11:309.

\bibitem[\protect\citeauthoryear{Lichtsteiner, Posch, and
  Delbruck}{2008}]{lichtsteiner2008128}
Lichtsteiner, P.; Posch, C.; and Delbruck, T.
\newblock 2008.
\newblock A 128$\times$128 120 db 15$\mu$s latency asynchronous temporal
  contrast vision sensor.
\newblock {\em IEEE Journal of Solid-State Circuits} 43(2):566--576.

\bibitem[\protect\citeauthoryear{Liu and Furber}{2016}]{liu2016noisy}
Liu, Q., and Furber, S.
\newblock 2016.
\newblock Noisy softplus: a biology inspired activation function.
\newblock In {\em International Conference on Neural Information Processing},
  405--412.
\newblock Springer.

\bibitem[\protect\citeauthoryear{Liu \bgroup et al\mbox.\egroup
  }{2019}]{liu2019unsupervised}
Liu, Q.; Pan, G.; Ruan, H.; Xing, D.; Xu, Q.; and Tang, H.
\newblock 2019.
\newblock Unsupervised aer object recognition based on multiscale
  spatio-temporal features and spiking neurons.
\newblock {\em arXiv preprint arXiv:1911.08261}.

\bibitem[\protect\citeauthoryear{Mohemmed \bgroup et al\mbox.\egroup
  }{2012}]{mohemmed2012span}
Mohemmed, A.; Schliebs, S.; Matsuda, S.; and Kasabov, N.
\newblock 2012.
\newblock Span: Spike pattern association neuron for learning spatio-temporal
  spike patterns.
\newblock {\em International journal of neural systems} 22(04):1250012.

\bibitem[\protect\citeauthoryear{Natarajan \bgroup et al\mbox.\egroup
  }{2008}]{natarajan2008encoding}
Natarajan, R.; Huys, Q.~J.; Dayan, P.; and Zemel, R.~S.
\newblock 2008.
\newblock Encoding and decoding spikes for dynamic stimuli.
\newblock {\em Neural computation} 20(9):2325--2360.

\bibitem[\protect\citeauthoryear{Neil and Liu}{2016}]{neil2016effective}
Neil, D., and Liu, S.-C.
\newblock 2016.
\newblock Effective sensor fusion with event-based sensors and deep network
  architectures.
\newblock In {\em 2016 IEEE International Symposium on Circuits and Systems
  (ISCAS)},  2282--2285.
\newblock IEEE.

\bibitem[\protect\citeauthoryear{Orchard \bgroup et al\mbox.\egroup
  }{2015a}]{orchard2015converting}
Orchard, G.; Jayawant, A.; Cohen, G.~K.; and Thakor, N.
\newblock 2015a.
\newblock Converting static image datasets to spiking neuromorphic datasets
  using saccades.
\newblock {\em Frontiers in Neuroscience} 9:437.

\bibitem[\protect\citeauthoryear{Orchard \bgroup et al\mbox.\egroup
  }{2015b}]{orchard2015hfirst}
Orchard, G.; Meyer, C.; Etienne-Cummings, R.; Posch, C.; Thakor, N.; and
  Benosman, R.
\newblock 2015b.
\newblock Hfirst: A temporal approach to object recognition.
\newblock {\em IEEE Transactions on Pattern Analysis and Machine Intelligence}
  37(10):2028--2040.

\bibitem[\protect\citeauthoryear{P{\'e}rez-Carrasco \bgroup et al\mbox.\egroup
  }{2013}]{perez2013mapping}
P{\'e}rez-Carrasco, J.~A.; Zhao, B.; Serrano, C.; Acha, B.;
  Serrano-Gotarredona, T.; Chen, S.; and Linares-Barranco, B.
\newblock 2013.
\newblock Mapping from frame-driven to frame-free event-driven vision systems
  by low-rate rate coding and coincidence processing--application to
  feedforward convnets.
\newblock {\em IEEE Transactions on Pattern Analysis and Machine Intelligence}
  35(11):2706--2719.

\bibitem[\protect\citeauthoryear{Ponulak and
  Kasi{\'n}ski}{2010}]{ponulak2010supervised}
Ponulak, F., and Kasi{\'n}ski, A.
\newblock 2010.
\newblock Supervised learning in spiking neural networks with resume: sequence
  learning, classification, and spike shifting.
\newblock {\em Neural Computation} 22(2):467--510.

\bibitem[\protect\citeauthoryear{Posch, Matolin, and
  Wohlgenannt}{2011}]{posch2011qvga}
Posch, C.; Matolin, D.; and Wohlgenannt, R.
\newblock 2011.
\newblock A qvga 143 db dynamic range frame-free pwm image sensor with lossless
  pixel-level video compression and time-domain cds.
\newblock {\em IEEE Journal of Solid-State Circuits} 46(1):259--275.

\bibitem[\protect\citeauthoryear{Qi \bgroup et al\mbox.\egroup
  }{2018}]{qi2018jointly}
Qi, Y.; Shen, J.; Wang, Y.; Tang, H.; Yu, H.; Wu, Z.; and Pan, G.
\newblock 2018.
\newblock Jointly learning network connections and link weights in spiking
  neural networks.
\newblock In {\em IJCAI},  1597--1603.

\bibitem[\protect\citeauthoryear{Riesenhuber and
  Poggio}{1999}]{riesenhuber1999hierarchical}
Riesenhuber, M., and Poggio, T.
\newblock 1999.
\newblock Hierarchical models of object recognition in cortex.
\newblock {\em Nature Neuroscience} 2(11):1019.

\bibitem[\protect\citeauthoryear{Serrano-Gotarredona and
  Linares-Barranco}{2015}]{serrano2015poker}
Serrano-Gotarredona, T., and Linares-Barranco, B.
\newblock 2015.
\newblock Poker-dvs and mnist-dvs. their history, how they were made, and other
  details.
\newblock {\em Frontiers in Neuroscience} 9:481.

\bibitem[\protect\citeauthoryear{Sironi \bgroup et al\mbox.\egroup
  }{2018}]{sironi2018hats}
Sironi, A.; Brambilla, M.; Bourdis, N.; Lagorce, X.; and Benosman, R.
\newblock 2018.
\newblock Hats: Histograms of averaged time surfaces for robust event-based
  object classification.
\newblock In {\em Proceedings of the IEEE Conference on Computer Vision and
  Pattern Recognition},  1731--1740.

\bibitem[\protect\citeauthoryear{Wu \bgroup et al\mbox.\egroup
  }{2014}]{wu2014cyborg}
Wu, Z.; Pan, G.; Principe, J.~C.; and Cichocki, A.
\newblock 2014.
\newblock Cyborg intelligence: Towards bio-machine intelligent systems.
\newblock {\em IEEE Intelligent Systems} 29(6):2--4.

\bibitem[\protect\citeauthoryear{Wu, Pan, and Zheng}{2013}]{wu2013cyborg}
Wu, Z.; Pan, G.; and Zheng, N.
\newblock 2013.
\newblock Cyborg intelligence.
\newblock {\em IEEE Intelligent Systems} 28(5):31--33.

\bibitem[\protect\citeauthoryear{Yu \bgroup et al\mbox.\egroup
  }{2013}]{yu2013precise}
Yu, Q.; Tang, H.; Tan, K.~C.; and Li, H.
\newblock 2013.
\newblock Precise-spike-driven synaptic plasticity: Learning hetero-association
  of spatiotemporal spike patterns.
\newblock {\em Plos one} 8(11):e78318.

\bibitem[\protect\citeauthoryear{Zhang \bgroup et al\mbox.\egroup
  }{2019}]{zhang2019mpd}
Zhang, M.; Wu, J.; Chua, Y.; Luo, X.; Pan, Z.; Liu, D.; and Li, H.
\newblock 2019.
\newblock Mpd-al: an efficient membrane potential driven aggregate-label
  learning algorithm for spiking neurons.
\newblock In {\em Proceedings of the AAAI Conference on Artificial
  Intelligence}, volume~33,  1327--1334.

\bibitem[\protect\citeauthoryear{Zhao \bgroup et al\mbox.\egroup
  }{2015}]{zhao2015feedforward}
Zhao, B.; Ding, R.; Chen, S.; Linares-Barranco, B.; and Tang, H.
\newblock 2015.
\newblock Feedforward categorization on aer motion events using cortex-like
  features in a spiking neural network.
\newblock {\em IEEE Transactions on Neural Networks and Learning Systems}
  26(9):1963--1978.

\end{thebibliography}

\end{document}